\documentclass[sigconf]{acmart}
\usepackage{multirow}
\usepackage{colortbl}
\usepackage{makecell}
\usepackage{pifont}
\usepackage{xcolor}
\usepackage[table]{xcolor}
\usepackage{enumitem}
\AtBeginDocument{%
  }

\copyrightyear{2026}
\acmYear{2026}
\setcopyright{cc}
\setcctype{by}
\acmConference[MM '26]{Proceedings of the 34th ACM International Conference on Multimedia}{November 10--14, 2026}{Rio de Janeiro, Brazil}
\acmBooktitle{Proceedings of the 34th ACM International Conference on Multimedia (MM '26), November 10--14, 2026, Rio de Janeiro, Brazil}
\acmDOI{10.1145/3767308.3835565}
\acmISBN{979-8-4007-2213-4/2026/11}

\begin{document}

\title{Semantic Steering for Controllable Generation: Tuning-Free Concept Erasure in Multimodal Diffusion Transformers}

\author{Qiao Li}
\orcid{0009-0004-1915-2570}
\affiliation{
  \institution{Institute of Information Engineering, Chinese Academy of Sciences}
  \city{Beijing}
  \country{China\\}
  \institution{School of Cyber Security, University of Chinese Academy of Sciences}
  \city{Beijing}
  \country{China}
}
\email{liqiao@iie.ac.cn}

\author{Xiaomeng Fu}
\orcid{0000-0001-7195-0765}
\affiliation{%
  \institution{Institute of Information Engineering, Chinese Academy of Sciences}
  \city{Beijing}
  \country{China\\}
  \institution{School of Cyber Security, University of Chinese Academy of Sciences}
  \city{Beijing}
  \country{China}
}
\email{fuxiaomeng@iie.ac.cn}

\author{Yuanshu Zhao}
\orcid{0009-0007-1103-6788}
\affiliation{%
  \institution{Institute of Information Engineering, Chinese Academy of Sciences}
  \city{Beijing}
  \country{China\\}
  \institution{School of Cyber Security, University of Chinese Academy of Sciences}
  \city{Beijing}
  \country{China}
}
\email{zhaoyuanshu@iie.ac.cn}

\author{Qipeng Wang}
\orcid{0009-0000-8297-7760}
\affiliation{%
  \institution{Institute of Information Engineering, Chinese Academy of Sciences}
  \city{Beijing}
  \country{China\\}
  \institution{School of Cyber Security, University of Chinese Academy of Sciences}
  \city{Beijing}
  \country{China}
}
\email{wangqipeng@iie.ac.cn}

\author{Jiao Dai}
\correspondingauthor
\orcid{0000-0003-3559-8009}
\affiliation{%
  \institution{Institute of Information Engineering, Chinese Academy of Sciences}
  \city{Beijing}
  \country{China}
}
\email{daijiao@iie.ac.cn}

\author{Jizhong Han}
\orcid{0000-0003-1107-3873}
\affiliation{%
  \institution{Institute of Information Engineering, Chinese Academy of Sciences}
  \city{Beijing}
  \country{China}
}
\email{hanjizhong@iie.ac.cn}
\renewcommand{\shortauthors}{Qiao Li et al.}

\begin{abstract}
  Multimodal Diffusion Transformers (MM-DiTs) have demonstrated remarkable text-to-image generation performance, surpassing traditional U-Net-based diffusion models. Nevertheless, their powerful generative capabilities also raise significant safety concerns, as they may generate sensitive or inappropriate content. While existing concept erasure methods aim to mitigate such risks, most require modifying model parameters, which are often architecture-specific and impractical for deployed larger models. Several tuning-free approaches face challenges when applied to advanced large-scale MM-DiTs due to their deeply embedded knowledge, broad semantic space, and context-dependent text encoders. To address these challenges, we propose to erase concepts by directly manipulating the model's internal representations. Our key insight, derived from an in‑depth analysis of MM‑DiT’s block‑wise generative roles, is that text-conditioned semantic representations are most salient in the middle blocks of MM-DiTs. Based on this, we extract representations of an unwanted concept and a desirable safe one from the middle block, construct a steering vector from their difference, and inject this single vector into consecutive early and middle blocks. By operating exclusively on the sparse text‑branch tokens and leveraging the straight sampling trajectory of rectified flow, our method achieves effective concept erasure with negligible overhead and without any training. Extensive experiments across MM-DiT models demonstrate that our method achieves state-of-the-art performance in erasing diverse concepts, enables effective control over the final output, and remains robust to adversarial attacks.
\end{abstract}

\ccsdesc[300]{Security and privacy~Usability in security and privacy}
\ccsdesc[300]{Computing methodologies~Computer vision}

\keywords{Multimodal Diffusion Transformers; Concept Erasure}
\begin{teaserfigure}
\centering
  \includegraphics[width=0.92\textwidth]{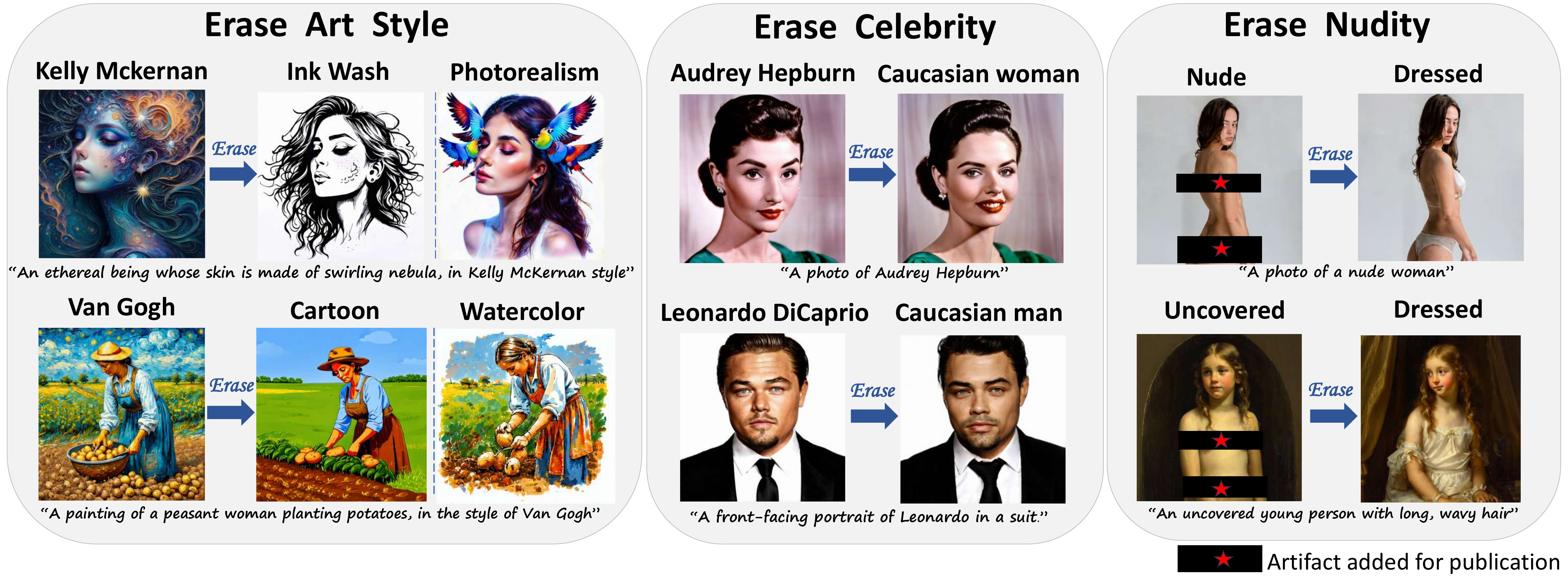}
  \caption{During the generation process with the multimodal diffusion transformer, our method applies steering vectors to erase diverse unwanted concepts, such as art styles, celebrities, and nudity, while supporting controllable generation.}
  \label{fig:teaser}
\end{teaserfigure}


\maketitle

\section{Introduction}
\label{sec:intro}
Recently, multimodal diffusion transformers (MM-DiTs)~\cite{Peebles2022ScalableDM}, such as Stable Diffusion 3 series~\cite{Esser2024ScalingRF} and FLUX series~\cite{blackforestlabs2024}, have emerged as the leading paradigm in text-to-image generation. These models demonstrate stronger multimodal understanding and superior generation quality at high resolutions, outperforming traditional U-Net-based diffusion models~\cite{Ho2020DenoisingDP,Song2019GenerativeMB,Song2020ScoreBasedGM,Rombach2021HighResolutionIS}. However, their remarkable generative capabilities also pose significant safety risks~\cite{Barez2025OpenPI,Wei2025ResponsibleDA}. Trained on large-scale web-crawled data, these models may inadvertently synthesize inappropriate or sensitive content, including not-safe-for-work (NSFW) material, copyrighted works, and images of real persons. To mitigate these risks, concept erasure techniques have been developed to prevent models from generating undesirable target concepts.

Most existing concept erasure methods~\cite{Zhang2023ForgetMeNotLT,Gandikota2023UnifiedCE,Kumari2023AblatingCI,Gandikota2023ErasingCF,Gong2024ReliableAE,Bui2024ErasingUC,Gao2024EraseAnythingEC,Fuchi2024ErasingCF,Li2025SetYS} require modifying the model parameters to remove or suppress unwanted concepts. However, many are architecture-specific and not directly transferable to more advanced MM-DiTs. Moreover, these methods may significantly degrade model's generative capabilities and typically impose substantial computational overhead, making them impractical for large-scale, real-world deployments.

To address these issues, several model tuning-free concept erasure methods~\cite{Schramowski2022SafeLD,Jain2024TraSCETS,Yoon2024SAFREETA,na2025trainingfree} have been proposed. Most of these methods typically leverage prompt-based interventions to guide the model away from predefined negative terms, demonstrating effectiveness in traditional U-Net-based diffusion models. 

Unfortunately, these tuning-free erasure methods face critical limitations when applied to MM-DiT. First, due to extensive pre-training on large-scale datasets, MM-DiT embeds target concepts within its deep representations, forming intrinsic and robust knowledge that is largely unresponsive to superficial prompt-based interventions. Second, MM-DiT has an expanded semantic space, making it difficult for negative prompts (which rely on explicit keyword specifications) to address the full range of concept variations and implicit associations, often leading to incomplete erasure. Third, some methods rely on linear separability of concepts in the token-level textual embedding space, which makes it challenging to apply them to sentence‑level, context‑dependent representations in the larger T5-XXL~\cite{Raffel2019ExploringTL} text encoder used in MM-DiT.

These limitations motivate a paradigm shift from external prompt engineering to direct intervention within the model's internal representations. Recent works in large language models~\cite{Dathathri2019PlugAP,Subramani2022ExtractingLS,Rimsky2023SteeringL2,Turner2023SteeringLM,Zhao2026ODESteerAU} and text-to-image models~\cite{Gaintseva2025CASteerCS,rodriguez2025controlling,DBLP:conf/icml/CywinskiD25} have demonstrated the feasibility of controlling model's output by steering internal activations at inference time. However, LLM‑based methods are tailored to the discrete, autoregressive generation paradigm of language, rendering them ill‑suited for the continuous, iterative denoising process and the fundamentally different dual-modality architecture of MM-DiT. Most existing T2I-based methods are designed for small-scale UNet architectures and either demand per-block/per-timestep interventions or extensive training, making them inherently inefficient or even infeasible for the deeper, multimodal blocks of MM-DiT architecture. As a result, steering internal representations for controllable generation in MM‑DiT remains largely unexplored.

To fill this gap, we investigate the internal generative mechanisms of MM‑DiT and propose a lightweight yet effective semantic steering method for controllable concept erasure, exploiting its intrinsic properties. We first conduct an in-depth analysis of the role of different multimodal blocks in image generation, revealing that middle blocks primarily encode the semantic representations of target concepts, while early and late blocks encode the overall image structures and fine-grained details, respectively. Such observation leads to a key insight: \textbf{we can extract highly salient semantic representations of a concept directly from the middle block of MM-DiT}. This naturally inspires us to leverage such representations to provide precise control over the model's output. Specifically, given a contrastive pair of an unwanted concept and a desirable safe concept, we extract their respective semantic representations from the middle block and construct a steering vector by computing their difference. We then leverage the distinct roles of early and middle blocks to inject the vector into key consecutive blocks, achieving controllable concept erasure while preserving overall image fidelity with a single vector. Unlike prior text-to-image intervention works that rely on image token representations, our method efficiently constructs and injects the steering vector purely on the sparser text-branch tokens, thereby avoiding the redundancy introduced by the large number of image tokens. The steering vector guides generation coherently across all denoising steps by following the straight sampling trajectory of rectified flow in MM-DiT, enabling effective erasure with negligible overhead.

Experiments on MM-DiT models demonstrate that our method achieves state-of-the-art performance in erasing diverse concepts, including celebrity, art style, and nudity, without requiring any training or fine-tuning, while enabling effective control over the output and exhibiting strong robustness to adversarial attacks.

Our contributions are summarized as:
\begin{itemize}[topsep=5pt, partopsep=0pt, itemsep=2pt, partopsep=0pt]
    \item  We propose a lightweight yet effective semantic steering method tailored to large-scale MM-DiT architecture for controllable concept erasure at inference time.
    
    \item  We design a steering framework that exploits MM-DiT’s properties, including block-wise roles in generation, information disparity in dual modalities, and rectified flow sampling, thereby achieving concept erasure while preserving image quality with a single vector and negligible overhead.

    \item Extensive experiments demonstrate that our method achieves state-of-the-art performance in erasing diverse concepts (celebrity, art style, and nudity) across MM-DiT models (SDv3.5 and FLUX.1), with controllable outputs and robustness against multiple adversarial attacks.

\end{itemize}

\section{Related Work}
\label{sec:related work}

\subsection{Multimodal Diffusion Transformers}
Traditional text-to-image diffusion models are based on UNet~\cite{Ronneberger2015UNetCN} architectures and incorporate self-attention and text-to-image cross-attention mechanisms. These models are trained to predict noise during denoising process, using noise schedulers such as DDPM~\cite{Ho2020DenoisingDP}.

Recently, multimodal diffusion transformers have brought about two paradigm shifts: architecture and training formulation. In terms of architecture, these models adopt Transformer-based backbones~\cite{Peebles2022ScalableDM} with multiple blocks and introduce cross-modal attention. For instance, models like Stable Diffusion 3 series~\cite{Esser2024ScalingRF} employ joint self-attention mechanisms to concatenate text and image tokens within the key, query, and value vectors of the Transformer architecture. They also integrate a T5-XXL~\cite{Raffel2019ExploringTL} text encoder and CLIP-based~\cite{Radford2021LearningTV} text encoders (CLIP-L/14 and OpenCLIP-bigG/14) to improve text-image alignment and cross-modal representation. In terms of training formulation, these new models leverage rectified flow~\cite{Lipman2022FlowMF,Liu2022FlowSA} with velocity predictions, directly connecting noise and data distributions via straight trajectories rather than curved paths in traditional diffusion models.

\subsection{Concept Erasure in Diffusion Models}
Concept erasure techniques aim to prevent models from generating undesirable concepts, such as not-safe-for-work (NSFW) content, copyrighted artworks, and personal portraits. Most existing methods~\cite{Zhang2023ForgetMeNotLT,Kumari2023AblatingCI,Lu2024MACEMC,Gong2024ReliableAE,Bui2024ErasingUC,Gao2024EraseAnythingEC,Li2025SetYS,Gandikota2023UnifiedCE} achieve concept erasure by updating the model's weights. For example, FMN~\cite{Zhang2023ForgetMeNotLT} forces the model to forget unwanted concepts by locating and suppressing their corresponding cross-attention maps through fine-tuning. MACE~\cite{Lu2024MACEMC} uses LoRA~\cite{Hu2021LoRALA} to fine-tune the cross-attention blocks for each concept and jointly optimizes the parameters across multiple concepts. UCE~\cite{Gandikota2023UnifiedCE} derives a closed-form edition of the weights in diffusion models without fine-tuning. However, most of these methods suffer from degraded generative capability and substantial computational overhead, and some methods are highly architecture-specific, making them difficult to extend to MM-DiT. To overcome these problems, several tuning-free erasure methods~\cite{Schramowski2022SafeLD,Jain2024TraSCETS,Yoon2024SAFREETA,na2025trainingfree} typically rely on prompt-based intervention to alter the model's denoising trajectories. For instance, Negative Prompting (NP) is a commonly used technique that replaces the unconditional input in classifier-free guidance with a negative prompt. Safe Latent Diffusion (SLD)~\cite{Schramowski2022SafeLD} extends NP by incorporating three noise predictions to exclude unwanted semantics during denoising. TraSCE~\cite{Jain2024TraSCETS} introduces a loss‑based guidance mechanism to enhance the flexibility of negative prompting. Nevertheless, current tuning-free concept erasure methods struggle to effectively erase concepts when applied to larger‑scale MM‑DiTs, which embed target concepts deeply and exhibit a broader semantic space.

\subsection{Internal Activation Intervention} Recent works have explored manipulating internal activations at inference time to control model behavior.  LLM-based methods~\cite{Dathathri2019PlugAP,Subramani2022ExtractingLS,Rimsky2023SteeringL2,Turner2023SteeringLM,Zhao2026ODESteerAU,Leong2023SelfDetoxifyingLM,Wang2023TrojanAA} construct steering vectors based on key activation values that encode human-interpretable features, and induce desired changes in the generated textual output when added to the forward passes of a frozen LLM. However, these approaches are tailored to the discrete, autoregressive nature of language modeling, rendering them incompatible with the continuous denoising process and multi-modal architecture of MM‑DiT. For T2I-based methods, CASteer~\cite{Gaintseva2025CASteerCS} constructs per‑block and per‑timestep steering vectors from cross‑attention activations. SAeUron~\cite{DBLP:conf/icml/CywinskiD25} trains additional sparse autoencoders to identify interpretable features for target manipulation. Both are primarily designed for small‑scale UNet-based architectures, and their significant computational overhead limits their scalability to deeper and larger-scale MM‑DiTs. ACT~\cite{rodriguez2025controlling} controls the model output by learning transport maps. Although it can be applied to DiT-based models, it still needs per‑block mapping learning, thereby incurring non‑negligible overhead. Moreover, these methods rely on image token representations, which may contain largely redundant information that hinders efficient and precise manipulation of the target concept.



\begin{figure}[t!]
    \centering
    \includegraphics[width=0.9\columnwidth]{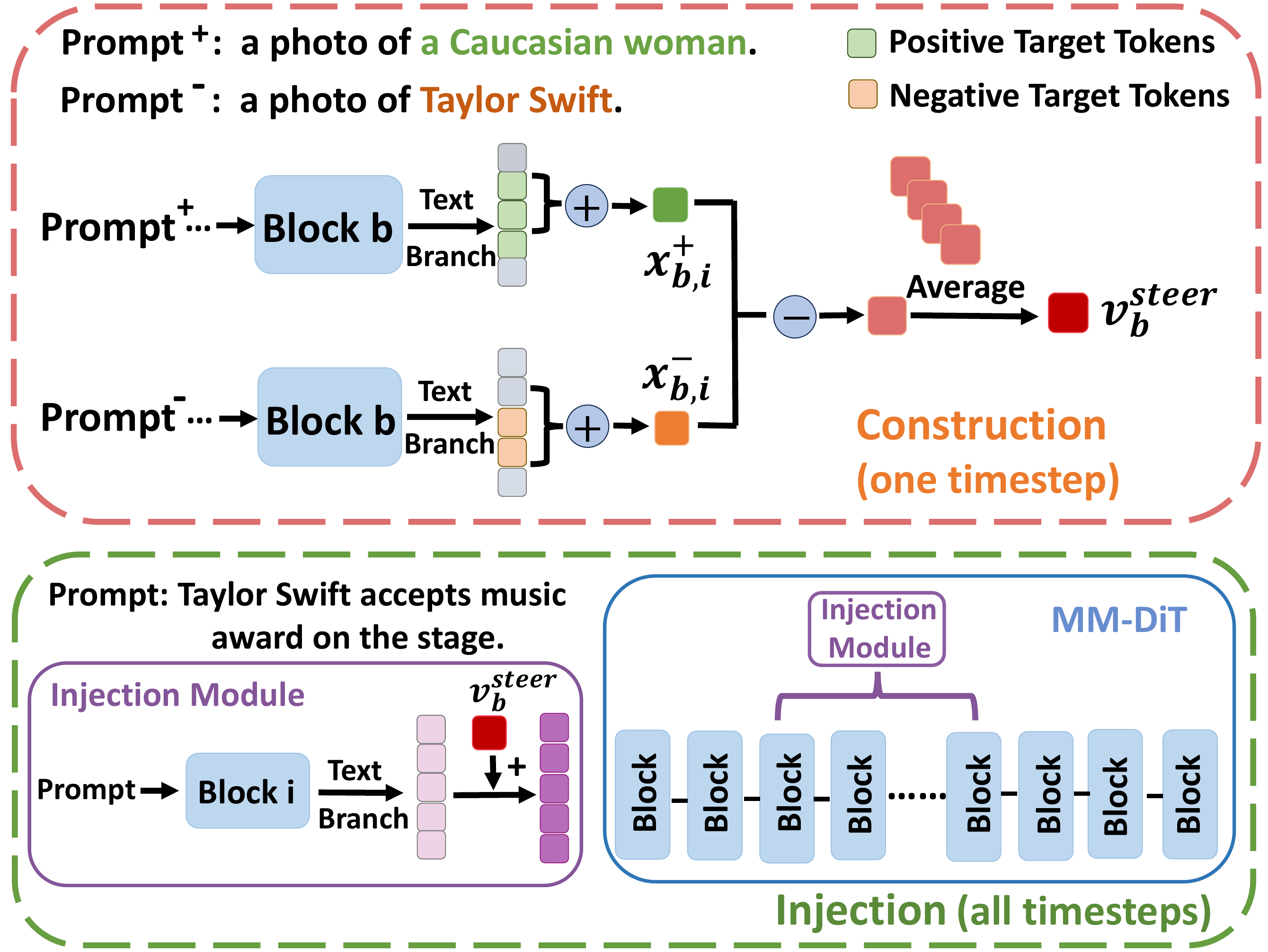} 
    \caption{The overall pipeline of our proposed method. We construct a steering vector from the output target tokens in the text branch of the middle block b \textcolor{magenta!50!black}{\cellcolor{magenta!10} \textbf{(top)}}. We then inject the vector into consecutive early and middle blocks \textcolor{green!50!black}{\cellcolor{green!10} \textbf{(bottom)}}. The construction process is performed at an intermediate denoising timestep, while the injection process is conducted coherently across all timesteps.}
    \label{fig:method}
\end{figure}

\section{Method}
Given an unwanted target concept and a desirable safe one, our goal is to construct a steering vector by computing the difference between their semantic representations, and inject it into key MM-DiT blocks to shift the unwanted output toward safe semantics. We first provide the preliminaries (Section~\ref{sec:preliminaries}), then analyze the role of various blocks and denoising timesteps in image generation, and construct a steering vector in the text branch accordingly (Section~\ref{sec:vector_construction}). Finally, we describe how the vector is injected spatially and temporally (Section~\ref{sec:steering}) to achieve effective target redirection. A pipeline of our method is illustrated in Figure~\ref{fig:method}.

\subsection{Preliminaries}
\label{sec:preliminaries}
We begin by analyzing MM-DiT's architecture. As a hierarchical model, it comprises numerous multimodal Transformer blocks. Taken Stable Diffusion-v3.5 as an example, it contains 24 Transformer blocks, and establishes a unified token space for visual and text modalities: $64 \times 64$ image tokens (flattened to 4096) in the image branch (when generating $1024\times1024$ images), and 154 or 333 text tokens (77 CLIP tokens plus 77 or 256 T5 tokens) in the text branch. Within each block, tokens interact via bidirectional cross-modal attention, with queries, keys, and values formed by concatenation: $Q=[Q_{text};Q_{image}]$, $K=[K_{text};K_{image}]$, and $V=[V_{text};V_{image}]$ This enables four interaction types: text-to-image (T2I), image-to-text (I2T), image-to-image (I2I), and text-to-text (T2T). The resulting representations from both modalities are output from block $b$ through separate branches and serve as input to block $b+1$, denoted as $X_{text,b}$ and $X_{image,b}$ (where $b \in \{1,…,B\}$).

Besides architecture, MM-DiT also differs significantly from traditional diffusion models in its sampling strategy, adopting rectified flow. Rectified flow models a transport map between two distributions $\pi_0$ (noise $N\sim(0, I)$) and $\pi_1$ (real data) by constructing straight-line trajectories between samples. To define these straight paths, the forward process is formulated as a linear interpolation in the latent space $D$:

\begin{figure}[t!]
    \centering
    \includegraphics[width=0.92\columnwidth]{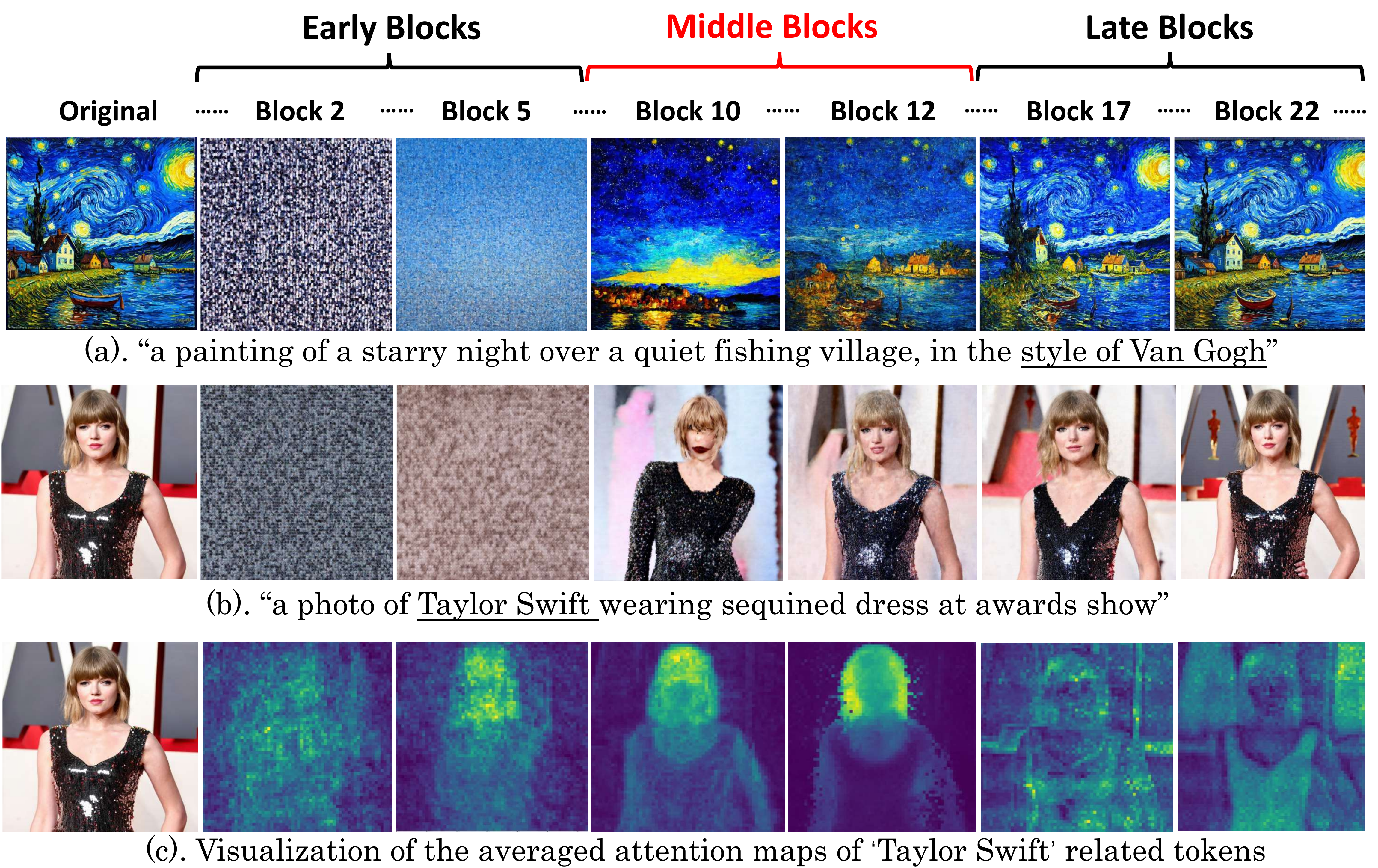} 
    \caption{(a) and (b) are images generated by SDv3.5, where random Gaussian noise is injected into distinct Transformer blocks (each image corresponds to noise injection in a specific block, with other blocks unchanged). (c) is a visualization of the averaged attention maps of target-related tokens.}
    \label{fig:add_noise_to_blocks}
\vspace{-5pt}
\end{figure}

\begin{equation}
    D_t=tD_1+(1-t)D_0,\>\>D_0\sim\pi_0,\>D_1\sim\pi_1
\end{equation}
The velocity field $V(D_t, t)$ is then trained to approximate
the dynamics of $D_t$. Once trained, it generates an image from a noise $Z$ across discrete steps from $t_T=0$ to $t_0=1$ by utilizing the velocity field to update $Z_t$ at each step:
\begin{equation}
    Z_{t_{i-1}}=Z_{t_i}+(t_{i-1}-t_i)V(Z_{t_i},t_i), \>\>i=T,…,1
\end{equation}

\subsection{Steering Vector Construction}
\label{sec:vector_construction}
To address the challenge of transferring existing tuning-free, prompt engineering-based concept erasure methods to MM-DiT, we seek to directly manipulate the model's internal representations to guide the generation process. To better understand the semantic distribution patterns within MM-DiT, we first analyze the role of different Transformer blocks, as well as denoising timesteps, in image generation. Based on the analysis, we then selectively leverage token representations from the text branch to construct a steering vector, which represents a shift direction to guide the unwanted concepts toward desired semantics.

\noindent \textbf{Semantic Focus in Middle Blocks.} To understand the distributions of primary semantic representations within MM-DiT, we first analyze how different blocks influence the image generation process. Specifically, we inject random Gaussian noise into the output features of intermediate blocks and observe the resulting image generation (examples in Figure~\ref{fig:add_noise_to_blocks} (a)-(b)). Noise in early blocks renders images completely unrecognizable due to global disruption, while noise in late blocks causes only minor fine-detail variations. Conversely, injecting noise into middle blocks significantly affects the formation of primary conceptual semantics (e.g., the swirling starry sky or Taylor Swift's face). These observations highlight the strong hierarchical characteristic of MM-DiT architecture: early blocks model global structure, late blocks refine surface details, and middle blocks are most expressive of core semantic content.

To further confirm that the middle blocks are the most representative for target semantics, we visualize the averaged T2I attention maps of target-related tokens (e.g., ``Taylor Swift'') across blocks (Figure~\ref{fig:add_noise_to_blocks} (c)). The attention patterns reveal that only in the middle blocks do such tokens attend significantly to the primary subject itself (e.g., Taylor’s face). These findings suggest that the middle blocks are the most suitable for extracting target concept's primary semantics. Conversely, capturing the primary semantics in the early or late blocks is ineffective, as target semantics are entangled with coarse global structures or overwhelming fine-grained details and thus difficult to extract.

\noindent \textbf{Balanced Representation at Intermediate Timesteps.} Besides analyzing the roles of model’s blocks, it is also crucial to consider the effect of denoising timesteps on semantic distributions. During the denoising process, MM-DiT adopts rectified flow sampling, where generations are iterative with representations evolving from coarse structures to fine details across timesteps. Similarly to the laws in MM-DiT blocks, early timesteps are dominated by global noise and contain limited semantics, whereas very late timesteps primarily encode minor details. Inspired by previous work~\cite{Meng2021SDEditGI}, which shows that features at intermediate timesteps preserve meaningful structural cues while retaining salient details and filtering out redundant ones, the intermediate denoising steps can provide a faithful representation of the target’s core semantics, effectively balancing structural coherence and key visual details.

\begin{table*}[]
\caption{Performance comparison between our method and baselines on erasing three conceptual categories from\textbf{ Stable Diffusion-v3.5-medium and FLUX.1}. Metrics are color-coded by their evaluation objective: \textcolor{yellow!50!black}{\cellcolor{yellow!50} \textbf{Yellow}} for erasure effectiveness, \textcolor{green!50!black}{\cellcolor{green!10} \textbf{Green}} for aesthetic quality post-erasure, and \textcolor{magenta!50!black}{\cellcolor{magenta!10} \textbf{Magenta}} for the impact on unrelated image quality (COCO-30K dataset). ↑ represents that a higher value indicates better performance, and vice versa. (Bold: best. Underline: second-best.)}
\large
\renewcommand{\arraystretch}{1.3}
\resizebox{0.95\linewidth}{!}{
\begin{tabular}{ccccccccccccccc}
\toprule[1.5pt]
\multirow{2}{*}{\textbf{Method}}  & \multicolumn{5}{c}{\textbf{Celebrity}}                            & \multicolumn{5}{c}{\textbf{Art Style}}               & \multicolumn{4}{c}{\textbf{Nudity}}              \\ 
 \cmidrule(lr){2-6} \cmidrule(lr){7-11} \cmidrule(lr){12-15}
                                  & \cellcolor{yellow!20} GIPHY↓         & \cellcolor{yellow!20} LLaVA↓         & \cellcolor{green!10} Aesthetic↑     & \cellcolor{magenta!10}FID↓   & \cellcolor{magenta!10} CLIP↑        &\cellcolor{yellow!20} Gram↓          &\cellcolor{yellow!20} LPIPS↑         & \cellcolor{green!10} Aesthetic↑ & \cellcolor{magenta!10} FID↓ &\cellcolor{magenta!10} CLIP↑ &\cellcolor{yellow!20} NudeNet↓       &\cellcolor{green!10} Aesthetic↑     &\cellcolor{magenta!10}  FID↓  &\cellcolor{magenta!10} CLIP↑         \\ \specialrule{0.9pt}{0.5ex}{0.5ex} 
\multicolumn{1}{l}{SDv3.5 (base)} & 0.602          & 0.524          & 5.621          & 17.85   &0.329       & 1              & 0              & 6.621      & 17.85 &0.329  & 0.739          & 5.706          & 17.85  &0.329        \\ 
SLD                               & \underline{0.029}          & \underline{0.022}          & 5.376          & 18.77  &\textbf{0.318}        & 0.271          & 0.318          & 6.474      & 18.99  &\underline{0.324}  & 0.526          & 5.716          & \underline{18.69}     &0.288     \\
TraSCE                            & 0.061          & 0.043          & 5.504          & \underline{18.42}    &0.299      & \underline{0.167}          & \underline{0.574}          & \textbf{6.608}      & 18.97 &0.320  & 0.460          & 5.366          & 18.81     &0.291     \\
STG           & 0.046          & 0.063          & \underline{5.511}         &18.79    &0.294      & 0.266          & 0.402         & \underline{6.601}      & 18.94 &0.312  & 0.490          & 5.799          & 18.78     &0.284
\\
NP                                & 0.034          & \underline{0.022}          & 5.414          & 18.49    &\underline{0.315}      & 0.180          & 0.536          & 6.442      & \textbf{18.66}  &0.318  & \underline{0.296}          & \underline{5.865}          & 18.77     & \underline{0.293}     \\ \rowcolor{blue!10}
\textbf{Ours}                     & \textbf{0.020} & \textbf{0.002} & \textbf{5.522} & \textbf{18.34}  &0.305  & \textbf{0.137} & \textbf{0.636} & 6.367      & \underline{18.91}  &\textbf{0.325}  & \textbf{0.220} & \textbf{5.868} & \textbf{18.60}   &\textbf{0.299} \\ \specialrule{0.9pt}{0.5ex}{0.5ex} 
FLUX.1 (base)           & 0.613          & 0.602          & 5.989          & 19.83  &0.319  & 1              & 0              & 6.832      & 19.83  &0.319  & 0.359          & 6.315          & 19.83  &0.319 \\ 
SLD                     & 0.433          & 0.398          & 5.918          &19.99   &\textbf{0.316}    & 0.234          & 0.708          & 6.626      & 20.34   &0.295   & 0.268          & 6.093          &       \textbf{19.97}  &0.308 \\
NP                      & 0.532          & 0.482          & 5.939          & 20.34   &0.307   & 0.217          & \underline{0.717}          & 6.653      & 21.02   &\underline{0.298}   & 0.364          & 6.189          &21.37    &\underline{0.315}   \\
STG           & 0.396          & 0.387          & 6.003         &20.15
&0.304      & 0.264          & 0.682         & 6.595      & 20.54 &0.290  & 0.248          & \underline{6.611}          & 20.53     &0.306
\\
UCE                     & 0.117          & 0.056          & 5.940          &\underline{19.94}  &0.301    & \underline{0.209}          & 0.714          & 6.587      &20.69   &\textbf{0.303}    & 0.167          & 6.124          & 20.98  &0.301    \\
ESD                     & \underline{0.059}          & \underline{0.005}          & 5.886          &21.15   &0.299    & 0.220          & 0.704          & 6.357      & \underline{20.33}   &0.283   & 0.239          & 6.606        &20.31      &0.298  \\
CA                      & 0.192          & 0.086          & \underline{6.133}          &20.78    &0.297   & 0.362          & 0.347          & \textbf{6.940}      & 20.97  &0.286    & \underline{0.044}          & 6.373          &      21.43  &0.304 \\ \rowcolor{blue!10}
\textbf{Ours}                    & \textbf{0.014} & \textbf{0.004} & \textbf{6.248} &\textbf{19.92}  &\underline{0.310}     & \textbf{0.192} & \textbf{0.729} & \underline{6.662}      &\textbf{20.01}  &0.294   & \textbf{0.023} & \textbf{6.631} &\underline{20.12}    &\textbf{0.316}   \\
\bottomrule[1.5pt]
\end{tabular}}
\label{tab:baselines}
\end{table*}

\noindent \textbf{Text Branch Vector Derivation.} The analysis on MM-DiT's blocks and denoising timesteps provide a key insight to achieve target concept erasure: primary concept semantics can be extracted most accurately in the middle blocks at intermediate timesteps. Inspired by prior works in LLMs~\cite{Dathathri2019PlugAP,Turner2023SteeringLM}, we can extract the representations of an unwanted concept along with a safe one. Their semantic difference can naturally serve as a steering vector encoding a safe shift direction. Our construction method begins with data preparation. Let $C^-$ be a prompt containing the target concept (e.g., a photo of Taylor Swift), and $C^+$ be an identical prompt where the target concept is replaced with a desired safe content (e.g., a photo of a Caucasian woman). For each target concept, we establish a dataset of n carefully curated paired prompts ($C_i^-$, $C_i^+$), where $i \in \{ 1,…,n \}$. This dataset is used to compute a steering vector by extracting the difference between the pairs. To derive the steering vector, we feed the prompt pairs into MM-DiT to obtain the output vectors from both modality branches in an intermediate block $b$: $X^-_{text,b}$, $X^+_{text,b}$ for text, and $X^-_{image,b}$, $X^+_{image,b}$ for image. Each vector lies in $\mathbb{R}^{N_{t} \times D_{f}}$, where $N_{t}$ is the number of tokens, and $D_{f}$ is the feature dimension. Given the overwhelming tokens in the image branch (4096 tokens), pinpointing critical tokens for difference extraction is difficult due to the large information redundancy. We therefore leverage the sparsity of the text branch (154 or 333 tokens), focusing exclusively on those textual tokens that differ between the paired prompts. The summed vectors of these targeted tokens for the i-th pair in block b are denoted as $x^-_{b,i}$ and $x^+_{b,i}$ (vectors in $\mathbb{R}^{D_{f}}$), which are used for the steering vector computation. We then compute the average difference between $x^+_{b,i}$ and $x^-_{b,i}$ for all the n pairs in block $b$ to derive a steering vector $v_b^{steer}$:
\begin{equation}
\label{eq:steer_vector_construction}
    v^{steer}_b=l \cdot \frac{v_b}{||v_b||}
\end{equation}
where:
\begin{equation}
    v_b=\frac{1}{n}\sum_{i=1}^{n} \Big(x_{b,i}^{+}-x_{b,i}^{-} \Big)
\end{equation}
The steering strength $l$ is determined empirically. The steering vector $v^{steer}_b \in \mathbb{R}^{D_{f}}$ defines the direction and magnitude for steering from a target concept to the erased safe content. 

\subsection{Steering Vector Injection}
\label{sec:steering}
Once a steering vector $v_b^{steer}$ is constructed, we add it to the output vectors of all text-encoded tokens in the key blocks of MM-DiT during generation, thereby steering the target concept's representations toward the desired safe one while preserving overall image quality. The strategy for injecting the steering vector is determined based on two key dimensions: (1) which intermediate blocks to target, and (2) which denoising timesteps to intervene.

\noindent \textbf{Progressive Injection into Specific Blocks.} A naive idea is to inject the vector directly into its construction block. However, this is proved ineffective for MM-DiT. This is because image features evolve progressively across blocks, and a small steering force in a single block is unlikely to alter the overall generation process.

\begin{figure*}[htbp]
    \centering
    \includegraphics[width=0.8\textwidth]{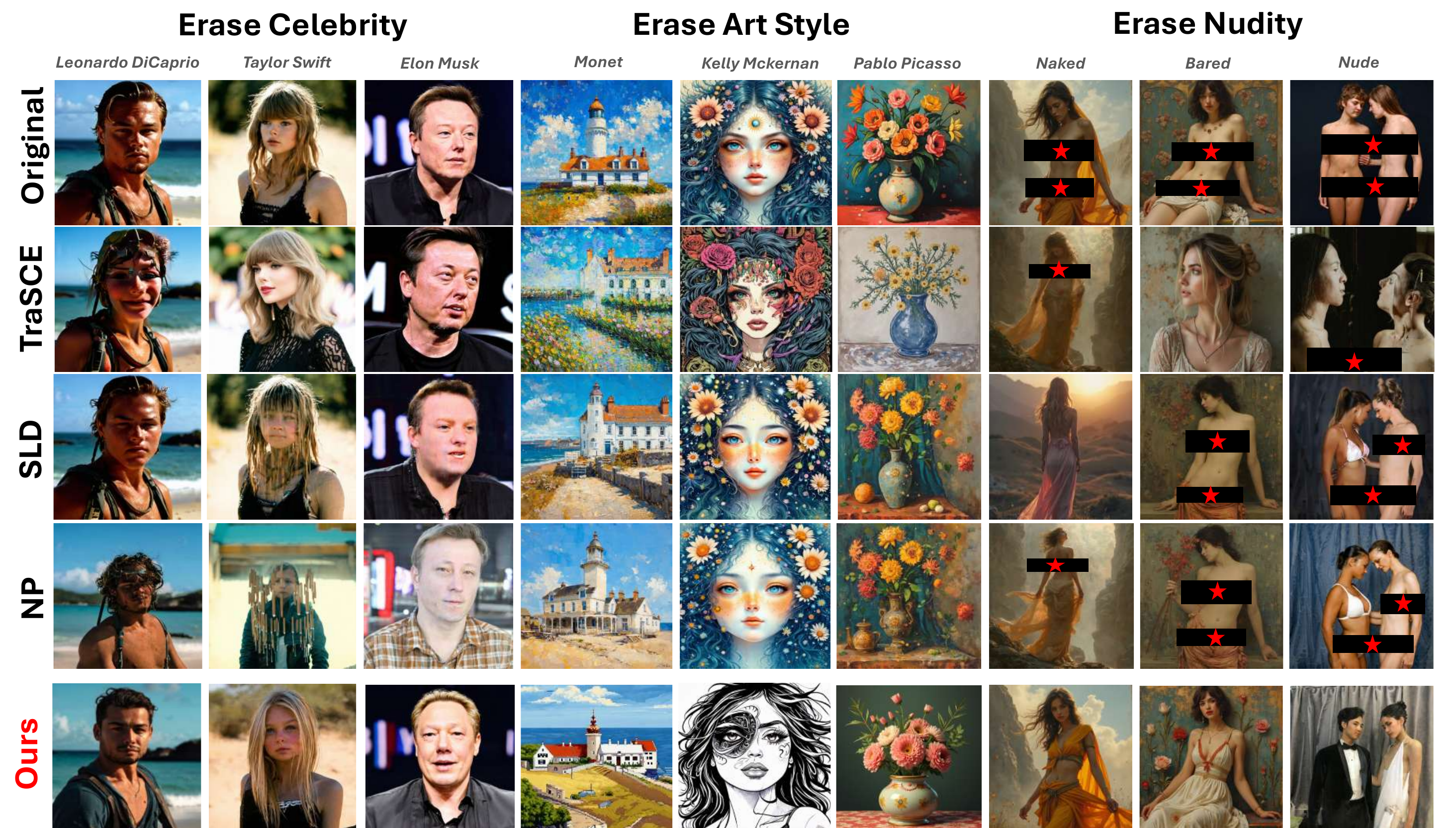} 
    \caption{Qualitative comparison between our method and baselines on Stable Diffusion-v3.5-medium. Our method effectively erases diverse concepts while largely preserving the overall image layout and quality. Specifically, for style erasure, our method erases Monet style to pixel art style, Kelly McKernan style to ink wash style, and Picasso style to photorealism style.}
    \label{fig:baselines}
\end{figure*}

To address this, we propose injecting the steering vector across multiple consecutive blocks, leveraging their cumulative effect for coherent steering. However, not all block combinations yield optimal results: (1) steering confined to middle or late blocks is ineffective, as the key structural and semantic basis is primarily established by early blocks, leaving limited room for subsequent alteration. (2) steering applied only to early blocks strongly influences image structure and may redirect the target semantics. Yet, the resulting structural changes often conflict with the subjects encoded by unchanged textual tokens in the middle blocks, causing distorted outputs or unintended semantic shifts.

To balance these trade-offs, we adopt a selective multi-block injection strategy: we inject the vector consecutively across multiple early and middle blocks. By gradually biasing feature representations in a unified direction, the steering signal accumulates across blocks, consistently guiding the model away from the original target while preserving image coherence. The very first few blocks, which primarily define the global layout without yet encoding main semantics, can be selectively omitted to avoid large distortion. Similarly, late blocks focusing on minor details have minimal impact on the overall semantic shift and can thus be selectively skipped.

\noindent \textbf{Stepwise Steering Across Denoising Timesteps.} Beyond selecting suitable blocks, the choice of denoising timestep plays a critical role in applying the steering vector. Unlike conventional diffusion models such as DDPM, which rely on a stochastic, nonlinear, and time-varying denoising trajectory, MM-DiT employs rectified flow, a deterministic generative process that follows a straight and stable sampling trajectory throughout generation. As a result, the features evolve in a more coherent and predictable manner, ensuring that a steering vector computed at any intermediate timestep captures a globally consistent semantic adjustment direction. This property enables the same vector to be applied consistently across all denoising steps, thereby ensuring temporally coherent and unified guidance throughout image generation process.

\section{Experiments}
\subsection{Experimental Setup}
\textbf{Model and Datasets.} We evaluate the effectiveness of our method on widely adopted multimodal diffusion transformer models, i.e., Stable Diffusion-v3.5-medium (SDv3.5)~\cite{Esser2024ScalingRF} and FLUX.1[DEV]~\cite{blackforestlabs2024} in the experiments. We target three conceptual categories: nudity, celebrity (Elon Musk, Taylor Swift, Leonardo DiCaprio, Steve Jobs, and Audrey Hepburn), and art styles (Van Gogh, Monet, Pablo Picasso, Kelly Mckernan, and Andy Warhol). We leverage GPT-4o~\cite{Achiam2023GPT4TR} to generate 100 test prompts for each concept. We utilize regular prompts from COCO-30K dataset~\cite{Lin2014MicrosoftCC} to assess the impact of our method on irrelevant image generation. For robustness evaluation, we use adversarial prompts from four datasets: Ring-A-Bell~\cite{Tsai2023RingABellHR}, MMA-Diffusion~\cite{Yang2023MMADiffusionMA}, I2P~\cite{Schramowski2022SafeLD}, and P4D~\cite{Chin2023Prompting4DebuggingRT}.

\noindent \textbf{Evaluation Metrics.} We evaluate our method from three aspects. First, to assess the erasure effectiveness, we employ GIPHY detector~\cite{GIPHYDetector} and the pretrained vision-language model LLaVA-1.5~\cite{Liu2023ImprovedBW} for celebrity detection, Gram matrix score~\cite{Gatys2015ANA} and LPIPS score~\cite{Zhang2018TheUE} for style evaluation, and NudeNet~\cite{bedapudi2019nudenet} for nudity detection. Second, to evaluate the image applicability after erasure, we use Aesthetic Predictor V2 Score~\cite{LAION-AES} for measuring visual appeal. Third, we use Frechet Inception Distance (FID)~\cite{Heusel2017GANsTB} and CLIP score~\cite{Radford2021LearningTV} for regular image quality evaluation.

\subsection{Comparison to Baselines}
We compare our methods with model tuning-free baselines that are applicable to SDv3.5 and FLUX.1: Safe Latent Diffusion (SLD)~\cite{Schramowski2022SafeLD}, TraSCE~\cite{Jain2024TraSCETS}, STG~\cite{na2025trainingfree}, and Negative Prompt (NP). We also choose model modification baselines that provide support for FLUX.1: UCE~\cite{Gandikota2023UnifiedCE}, ESD~\cite{Gandikota2023ErasingCF}, and CA~\cite{Kumari2023AblatingCI}. Quantitative and qualitative results are shown in Table~\ref{tab:baselines} and Figure~\ref{fig:baselines}, respectively. 

For celebrity and nudity erasure, our method outperforms all the baselines on the two models in both erasure effectiveness and aesthetic score, indicating that it precisely steers target concepts to desired safe content without compromising image quality. Besides, our method achieves nearly the lowest FID score among all baselines, demonstrating minimal impact on normal image generation.

For art style erasure, our method achieves the best erasure performance on both models, as validated by Gram matrix and LPIPS score. The aesthetic score is slightly lower than that of some baselines. We note, however, that this does not necessarily indicate a degradation in perceptual image quality. MM-DiTs inherently generate high-aesthetic images, and baselines with weaker erasure efficacy produce results that are visually similar to the original, thereby retaining high aesthetic scores. In contrast, our method performs a more complete style transformation, sometimes toward styles that are underrepresented in the predictor's training set, which may lower the score. Therefore, a slight drop in aesthetic score is an acceptable trade‑off for achieving thorough style removal.

{\Large
\begin{table}[t!]
\caption{Quantitative results of constructing steering vectors in different blocks or at different denoising timesteps.}
{\Large
\centering
\renewcommand{\arraystretch}{1.2}
\resizebox{0.9\columnwidth}{!}{
\begin{tabular}{ccccc}
\toprule[1.5pt]
\multirow{2}{*}{\textbf{Construction Method}}  & \multicolumn{2}{c}{\textbf{Celebrity}} & \multicolumn{2}{c}{\textbf{Art Style}}\\ 
\cmidrule(lr){2-3} \cmidrule(lr){4-5} 
                                      & GIPHY↓      & Aesthetic↑      & Gram↓       & Aesthetic↑  \\ \specialrule{0.8pt}{0.5ex}{0.5ex}
Early block                           & 0.002       & 5.053           & 0.186       & 6.287          \\
Late block                            & 0.025       & 5.384           & 0.402       & 6.452            \\ \rowcolor{blue!10}
\textbf{Ours (middle block)}          & 0.020       & 5.522           & 0.137       & 6.367            \\ \hline
Early timestep                        & 0.001       & 5.436           & 0.162       & 6.416           \\
Late timestep                         & 0.028       & 5.331           & 0.203       & 6.474        \\ \rowcolor{blue!10}
\textbf{Ours (intermediate timestep)} & 0.020       & 5.522           & 0.137       & 6.367           \\ \bottomrule[1.5pt]
\end{tabular}}}
\label{tab:block_construction}
\vspace{-7pt}
\end{table}}

\begin{figure}[t!]
    \centering
    \includegraphics[width=0.94\columnwidth]{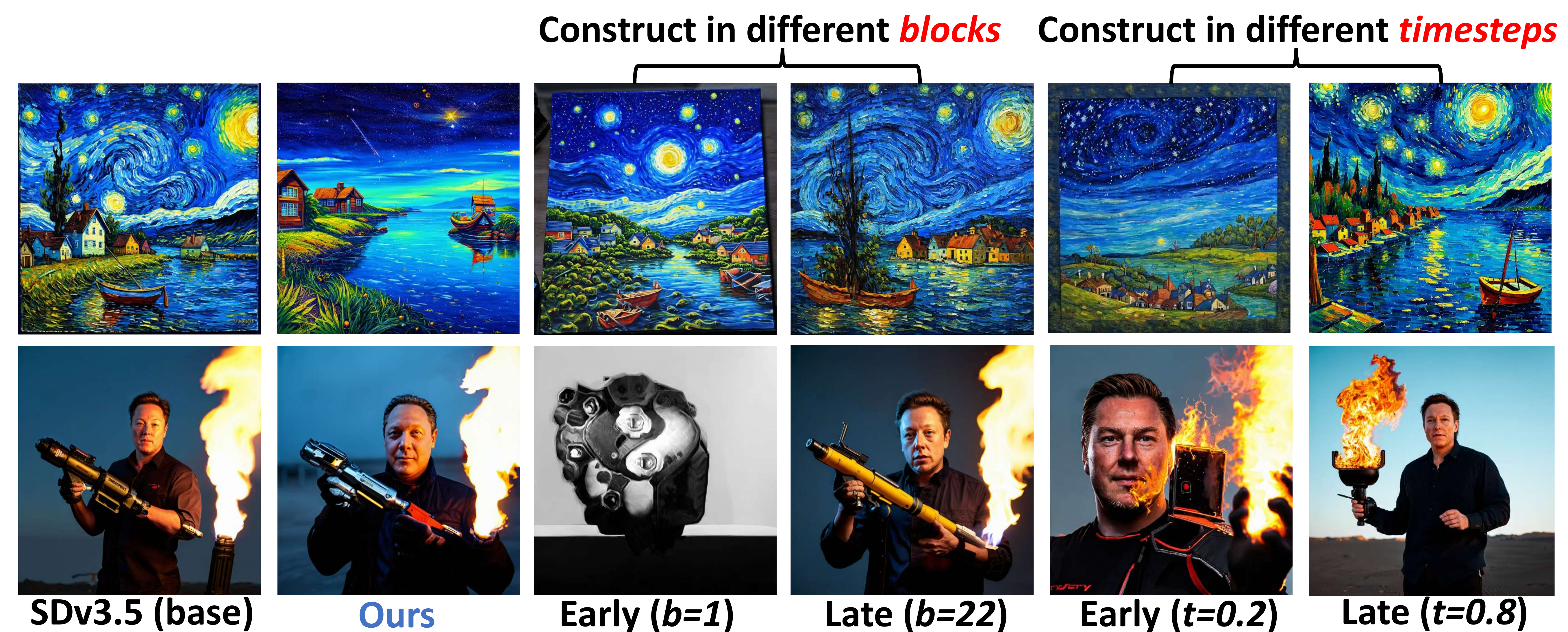} 
    \caption{Images generated using steering vectors constructed in different Transformer blocks or at different denoising timesteps. (Top: Van Gogh style. Bottom: Elon Musk).}
    \label{fig:block_construction}
\end{figure}

\subsection{Ablation Study}
To verify the design choices of our method, we conduct ablation studies for celebrity and art style erasure on SDv3.5.

\noindent \textbf{Block Selection for Steering Vector Construction.} In our method, we construct a steering vector in the middle block ($b=10$). To validate the block selection, we show the averaged results of constructing in the early blocks ($b=1, 3, 5, 7$) and the late blocks ($b=16, 18, 20, 22$) in Table~\ref{tab:block_construction}, and the example images after steering in Figure~\ref{fig:block_construction}. The vectors built in early blocks mainly capture coarse and global structural features. Consequently, applying these vectors steers or even distorts the overall image layout, leading to the disappearance of the target concept with a very low aesthetic score (low image quality). In contrast, the vectors built in late blocks primarily learn minor and even redundant details. As a result, their injection has negligible impact on the generated images. We note that $b=10$ is not the only effective choice, the interval $b\in[9,13]$ constitutes a stable region where the method remains effective.

\noindent \textbf{Timestep Selection for Steering Vector Construction.} During the denoising process, we construct a steering vector at an intermediate timestep ($t=0.5$). We assess the averaged results of vector construction at early ($t=0.1, 0.2, 0.3$) and late timesteps ($t=0.7, 0.8, 0.9$) respectively, shown in Table~\ref{tab:block_construction} and Figure~\ref{fig:block_construction}. Results validate that construction at early or late timesteps can also erase the concept to some extent, but it leads to significant alterations in overall image structure or detailed content.

{\Large
\begin{table}[t!]
\caption{Quantitative results of injecting steering vectors into different Transformer block combinations.}
\centering
\renewcommand{\arraystretch}{1.2}
\resizebox{0.92\columnwidth}{!}{
\begin{tabular}{ccccccc}
\toprule[1.5pt]
\multicolumn{3}{c}{\textbf{Injection Block Combinations}}          & \multicolumn{2}{c}{\textbf{Celebrity}} & \multicolumn{2}{c}{\textbf{Art Style}} \\ 
\cmidrule(lr){1-3} \cmidrule(lr){4-5} \cmidrule(lr){6-7} 
Single            & Multiple          & Position          & GIPHY↓       & Aesthetic↑       & Gram↓        & Aesthetic↑       \\ \specialrule{0.8pt}{0.5ex}{0.5ex}
{\textcolor{green!70!black}{\ding{51}}}           &                   & Early             & 0.395       & 5.532           & 0.252       & 6.688  \\
{\textcolor{green!70!black}{\ding{51}}}           &                   & Middle            & 0.625       & 5.522           & 0.541       & 6.635        \\
{\textcolor{green!70!black}{\ding{51}}}           &                   & Late              & 0.686       & 5.503           & 0.882       & 6.613      \\
                  & {\textcolor{green!70!black}{\ding{51}}}           & Early             & 0.322       & 5.268           & 0.115       & 5.006       \\
                  & {\textcolor{green!70!black}{\ding{51}}}           & Middle            & 0.547       & 5.537           & 0.351       & 6.597         \\
                  & {\textcolor{green!70!black}{\ding{51}}}           & Late              & 0.647       & 5.498           & 0.540       & 6.570          \\ \rowcolor{blue!10}
\multicolumn{3}{c}{\textbf{Ours (multiple early+middle)}} & 0.020       & 5.522           & 0.137       & 6.367         \\ \hline
\multicolumn{3}{c}{Block-wise local steering} & 0.029       & 5.341           & 0.158       & 6.177         \\
\bottomrule[1.5pt]
\end{tabular}}
\label{tab:injection_block}
\vspace{-5pt}
\end{table}}

\begin{figure}[t!]
    \centering
    \includegraphics[width=0.95\columnwidth]{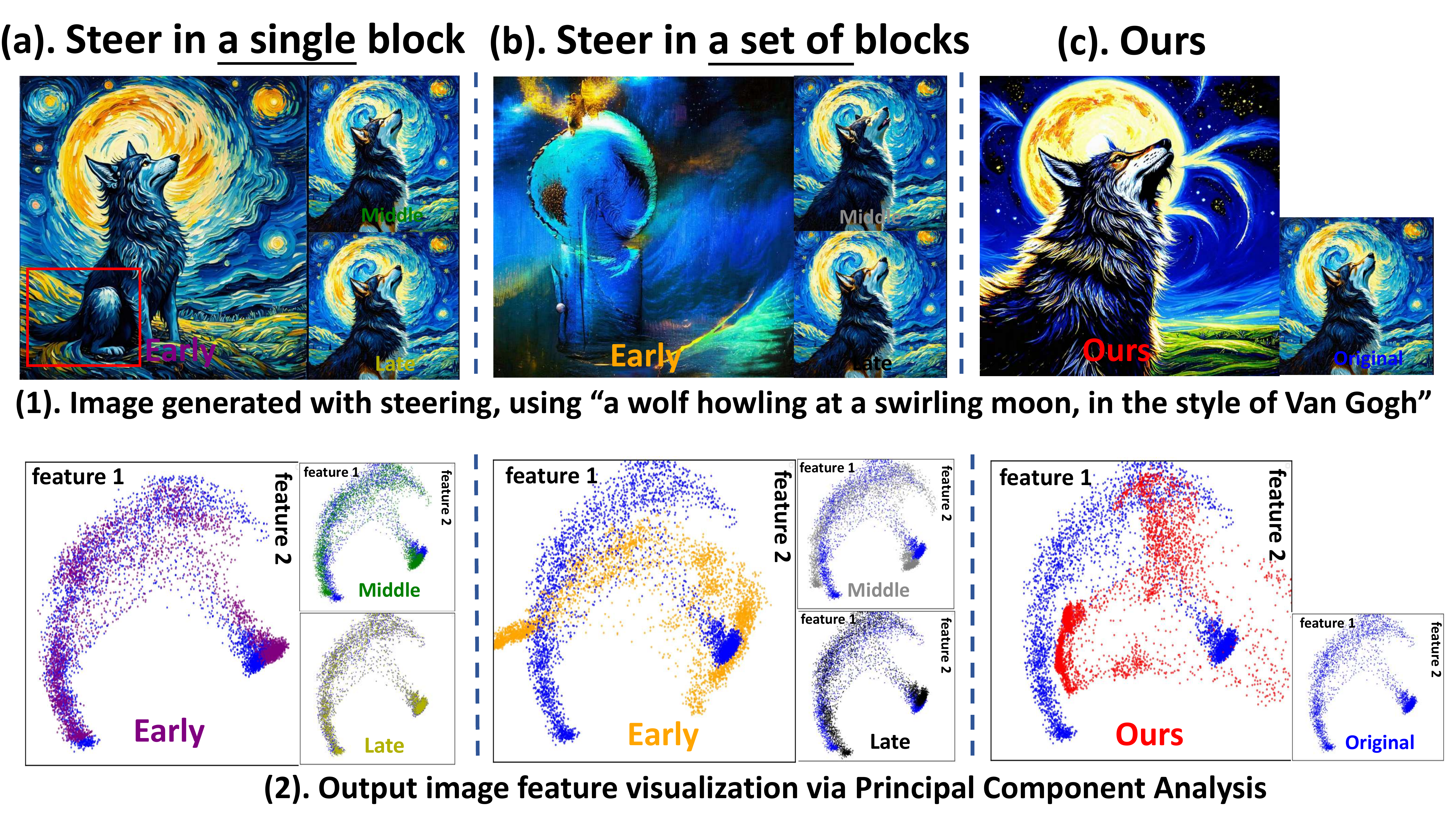} 
    \caption{Images generated by injecting vectors into different block combinations, along with the respective feature visualizations. The feature distribution after steering (\textcolor{red}{\cellcolor{red!30}\textbf{red}} points in 'Ours') shifts away from the original style distribution (\textcolor{blue}{\cellcolor{blue!30} \textbf{blue}} points in 'Original'), yet retains a similar overall structure.}
    \label{fig:feature_visualization}
\end{figure}

\noindent \textbf{Block Combinations for Steering Vector Injection.} Our method injects the steering vector into consecutive early and middle blocks ($b=3, …, 10$). As shown in Table~\ref{tab:injection_block}, we provide quantitative results of injecting into a single early, middle or late block, as well as into multiple early, middle, or late blocks. We also show the generated image examples and visualize their corresponding feature distributions via Principal Component Analysis (PCA) in Figure~\ref{fig:feature_visualization}. Results reveal that injecting into a single or multiple middle/late blocks causes nearly no changes, whereas injecting into a single early block may bring minor structural modifications. However, injecting into multiple early blocks leads to substantial, undesirable image distortion. We conclude that early blocks are crucial for overall image structure, while both early and middle blocks are critical for coherently constructing the main semantics of the image.

\noindent \textbf{Block-wise Local Steering.} To further analyze the alignment of steering directions across blocks, we independently construct and inject a local steering vector for each block $b\in[1, 10]$. As shown in the last row of Table~\ref{tab:injection_block}, the erasure effectiveness of this per‑block local steering is close to that of our method, indicating that the optimal steering directions are highly consistent across blocks and that a vector extracted from the middle block can effectively steer multiple blocks. However, multiple local vectors slightly reduce aesthetic quality, likely due to incoherent interventions that introduce subtle artifacts, while also increasing computational cost. Therefore, we choose a single middle vector instead of multiple local ones.

\begin{figure}[t!]
    \centering
    \includegraphics[width=0.95\columnwidth]{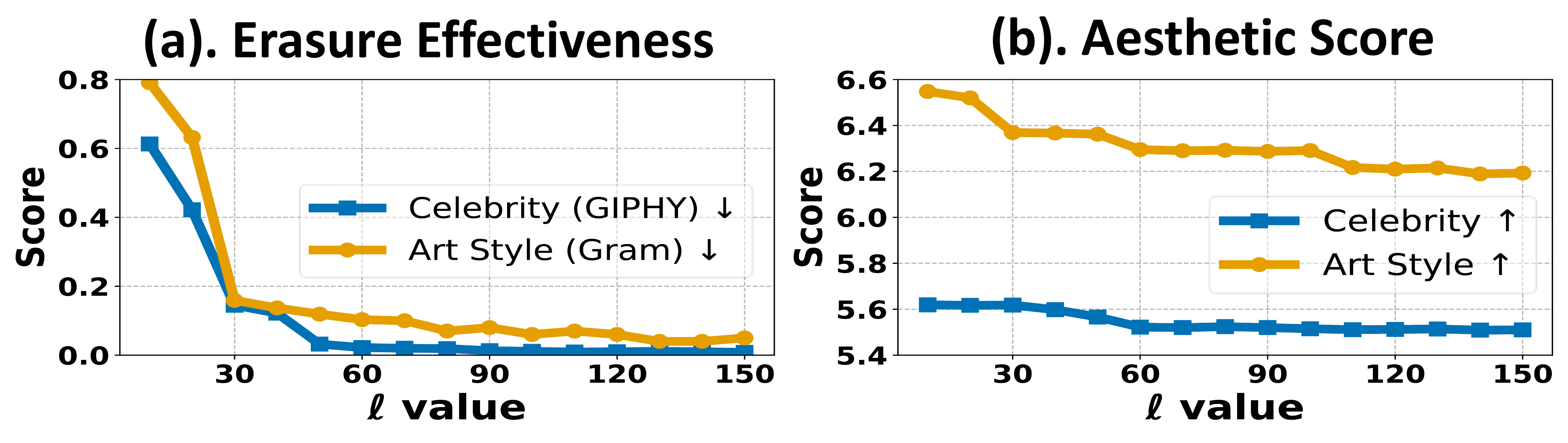} 
    \caption{Sensitivity analysis of the steering strength $\ell$. We evaluate how erasure effectiveness (GIPHY and Gram) and post-erasure image aesthetics vary with $\ell$.}
    \label{fig:l_curve}
    \vspace{-7pt}
\end{figure}

\begin{figure}[t!]
    \centering
    \includegraphics[width=0.92\columnwidth]{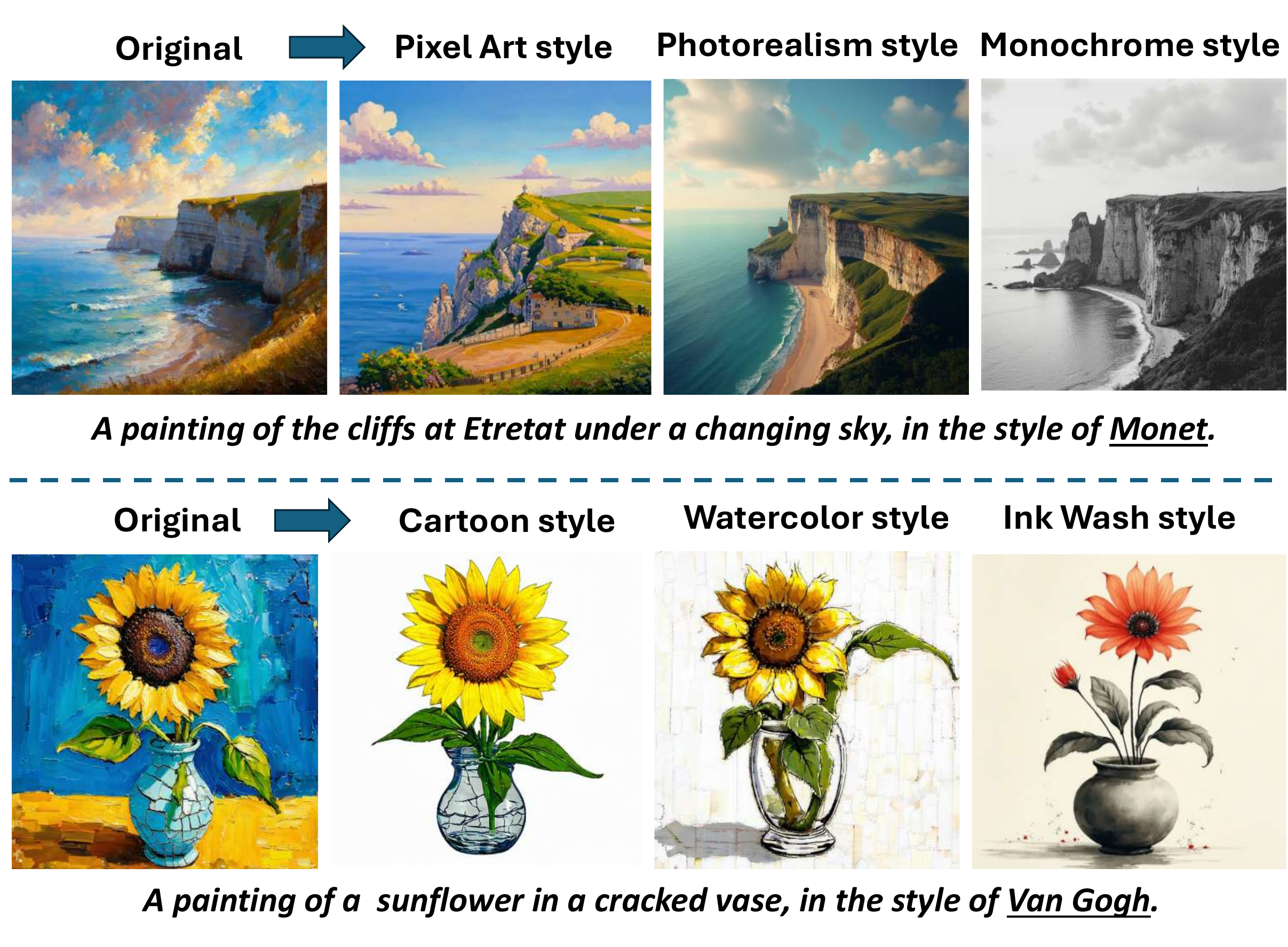} 
    \caption{Example images show multi-style steering for controlled style transformation: Monet style is steered into pixel art, photorealism, and monochrome, while Van Gogh style is steered into cartoon, watercolor, and ink wash styles.}
    \label{fig:multi_style}
\end{figure}

\noindent \textbf{Sensitivity Analysis of the Steering Strength $l$.} We present sensitivity analysis of $\l$ in Figure~\ref{fig:l_curve}. For celebrity, the erasure effectiveness increases sharply between $l=10$ and $l=60$, reaching a stable low value thereafter, while the aesthetic score decreases slightly and then plateaus. For art style, effectiveness improves rapidly up to $l=30$, after which the gain slows. The aesthetic score first declines, stabilizes between $l=30$ and $l=50$, and then drops again beyond $l=50$, likely because style information is globally distributed and is more susceptible to stronger steering than celebrity features. Based on this trade-off, we select $l=70$ for celebrity and $l=40$ for art style to balance erasure and aesthetic quality. Notably, for both concepts, there exists a stable plateau ($l\in[30,60]$ for art style and $l\in[60,100]$ for celebrity) where the method consistently removes the target while preserving image quality. This indicates the robustness of our method to the choice of $l$.

{\Large
\begin{table}[t!]
\caption{Robustness comparison between our method and baselines against four types of adversarial attacks targeting nudity on FLUX.1. All presented values are NudeNet scores.}
\centering
\renewcommand{\arraystretch}{1.1}
\resizebox{0.9\columnwidth}{!}{
\begin{tabular}{ccccc}
\toprule[1.5pt]
\textbf{Method}        & \textbf{Ring-A-Bell↓}   & \textbf{I2P↓}           & \textbf{MMA-Diffusion↓} & \textbf{P4D↓}   \\ \specialrule{0.8pt}{0.5ex}{0.5ex} 
FLUX.1 (base) & 0.435         & 0.148         & 0.075          & 0.238 \\ \hline
SLD           & 0.495         & 0.151         & 0.056          & 0.227 \\
NP            & 0.461         & 0.160        & 0.117         & 0.229 \\
STG           & 0.482         & 0.154        & 0.148         & 0.234 \\
UCE           & 0.415         & 0.147         & 0.086          & 0.187 \\
ESD           & \underline{0.127}         & \underline{0.065}          & 0.021          & \underline{0.071}  \\
CA            & 0.212         & 0.104         & \underline{0.020}          & 0.130 \\ \rowcolor{blue!10}
\textbf{Ours} & \textbf{0.057} & \textbf{0.014} & \textbf{0.016} & \textbf{0.013}\  \\ 
\bottomrule[1.5pt]
\end{tabular}}
\label{tab:robust}
\vspace{-5pt}
\end{table}}

\begin{figure}[t!]
    \centering
    \includegraphics[width=0.94\columnwidth]{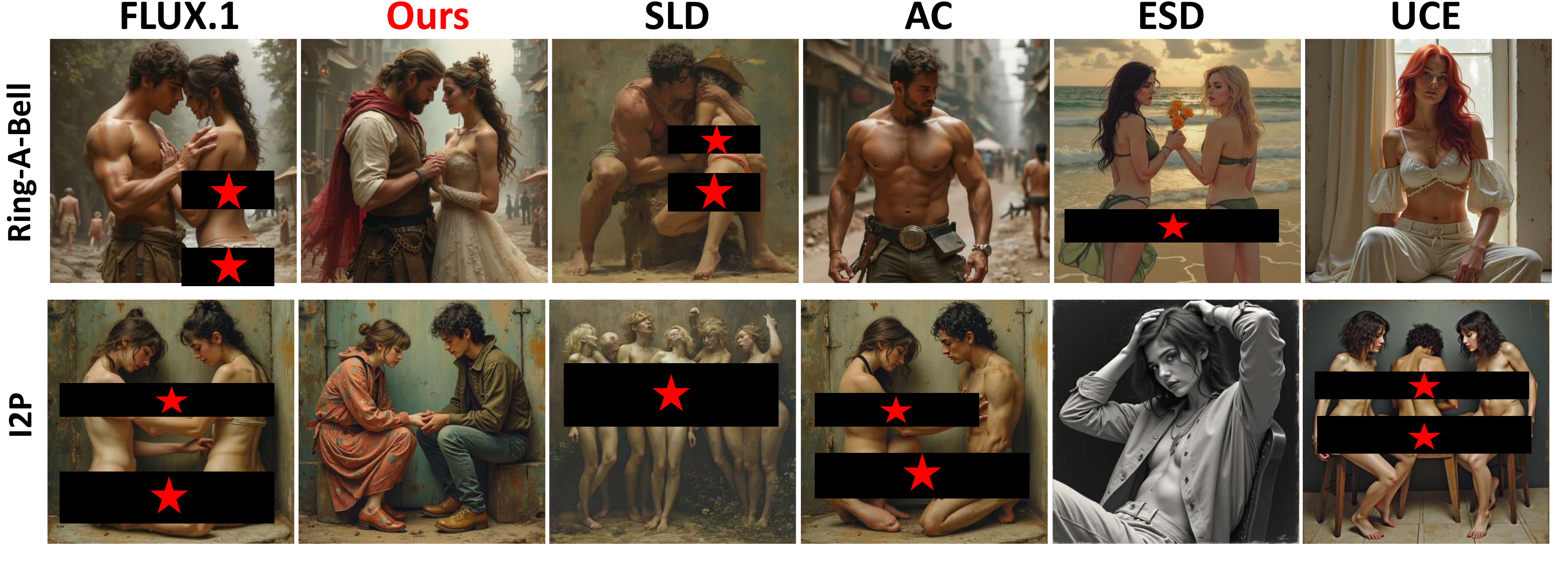} 
    \caption{Visual examples of nudity erasure against adversarial prompts generated by Ring-A-Bell and I2P. Our method exhibits strong robustness against adversarial prompts.}
    \label{fig:robust}
\end{figure}

\subsection{Multi-Style Steering}
While existing baseline methods primarily focus on removing the target concept without specifying the final output after erasure, our method supports more flexible steering—redirecting the target concept's representation toward diverse, specified semantic distributions. For example, as shown in Figure~\ref{fig:multi_style}, our method can transform an art style into different ones. This flexibility stems from style-specific steering vectors, each guiding the target concept’s representation toward a predefined output distribution. Applying the corresponding vectors allows users to achieve both target concept erasure and fine-grained control over desired semantic outcomes. 

\subsection{Robustness to Adversarial Attack}
We evaluate the robustness of our method against four adversarial attacks. Since certain attacks can only generate adversarial prompts targeting the 'nudity' concept, our analysis focuses on these attacks. As illustrated in the FLUX.1 results in Table~\ref{tab:robust} and Figure~\ref{fig:robust}, compared to the baselines, our approach exhibits the strongest defense performance across all tested attack types, obtaining the lowest NudeNet scores. This suggests that our method significantly reduces the proportion of generated nudity-related concepts.

\section{Conclusion}
In this paper, we investigate the intrinsic properties of MM-DiT and propose a tuning-free concept erasure framework. By analyzing the generative roles of different blocks, we reveal that semantic representations of target concepts are predominantly concentrated in the middle blocks. Building on this insight, we construct a steering vector that guides the model toward safe semantics, and inject it into key blocks at every denoising timestep, thereby achieving concept erasure while preserving the overall image quality. Our method not only improves generation safety but also provides a foundation for future research on controllable image generation.

\section{Acknowledgments}
This work was supported by the National Key Research and Development Program of China (No.2024YFC3307402).

\bibliographystyle{ACM-Reference-Format}
\balance
\bibliography{acmart}

@inproceedings{Liu2023ImprovedBW,
  title={Improved baselines with visual instruction tuning},
  author={Liu, Haotian and Li, Chunyuan and Li, Yuheng and Lee, Yong Jae},
  booktitle={Proceedings of the IEEE/CVF conference on computer vision and pattern recognition},
  pages={26296--26306},
  year={2024}
}

@inproceedings{Zhang2018TheUE,
  title={The unreasonable effectiveness of deep features as a perceptual metric},
  author={Zhang, Richard and Isola, Phillip and Efros, Alexei A and Shechtman, Eli and Wang, Oliver},
  booktitle={Proceedings of the IEEE conference on computer vision and pattern recognition},
  pages={586--595},
  year={2018}
}

@article{Heusel2017GANsTB,
  title={Gans trained by a two time-scale update rule converge to a local nash equilibrium},
  author={Heusel, Martin and Ramsauer, Hubert and Unterthiner, Thomas and Nessler, Bernhard and Hochreiter, Sepp},
  journal={Advances in neural information processing systems},
  volume={30},
  year={2017}
}

@inproceedings{Schramowski2022SafeLD,
  title={Safe latent diffusion: Mitigating inappropriate degeneration in diffusion models},
  author={Schramowski, Patrick and Brack, Manuel and Deiseroth, Bj{\"o}rn and Kersting, Kristian},
  booktitle={Proceedings of the IEEE/CVF conference on computer vision and pattern recognition},
  pages={22522--22531},
  year={2023}
}

@article{Jain2024TraSCETS,
  title={Trasce: Trajectory steering for concept erasure},
  author={Jain, Anubhav and Kobayashi, Yuya and Shibuya, Takashi and Takida, Yuhta and Memon, Nasir and Togelius, Julian and Mitsufuji, Yuki},
  journal={arXiv preprint arXiv:2412.07658},
  year={2024}
}

@inproceedings{Lin2014MicrosoftCC,
  title={Microsoft coco: Common objects in context},
  author={Lin, Tsung-Yi and Maire, Michael and Belongie, Serge and Hays, James and Perona, Pietro and Ramanan, Deva and Doll{\'a}r, Piotr and Zitnick, C Lawrence},
  booktitle={European conference on computer vision},
  pages={740--755},
  year={2014},
  organization={Springer}
}

@article{Ho2020DenoisingDP,
  title={Denoising diffusion probabilistic models},
  author={Ho, Jonathan and Jain, Ajay and Abbeel, Pieter},
  journal={Advances in neural information processing systems},
  volume={33},
  pages={6840--6851},
  year={2020}
}

@inproceedings{Rombach2021HighResolutionIS,
  title={High-resolution image synthesis with latent diffusion models},
  author={Rombach, Robin and Blattmann, Andreas and Lorenz, Dominik and Esser, Patrick and Ommer, Bj{\"o}rn},
  booktitle={Proceedings of the IEEE/CVF conference on computer vision and pattern recognition},
  pages={10684--10695},
  year={2022}
}

@article{Song2019GenerativeMB,
  title={Generative modeling by estimating gradients of the data distribution},
  author={Song, Yang and Ermon, Stefano},
  journal={Advances in neural information processing systems},
  volume={32},
  year={2019}
}

@article{Song2020ScoreBasedGM,
  title={Score-based generative modeling through stochastic differential equations},
  author={Song, Yang and Sohl-Dickstein, Jascha and Kingma, Diederik P and Kumar, Abhishek and Ermon, Stefano and Poole, Ben},
  journal={arXiv preprint arXiv:2011.13456},
  year={2020}
}

@inproceedings{Zhang2023ForgetMeNotLT,
  title={Forget-me-not: Learning to forget in text-to-image diffusion models},
  author={Zhang, Gong and Wang, Kai and Xu, Xingqian and Wang, Zhangyang and Shi, Humphrey},
  booktitle={Proceedings of the IEEE/CVF conference on computer vision and pattern recognition},
  pages={1755--1764},
  year={2024}
}

@inproceedings{Gandikota2023UnifiedCE,
  title={Unified concept editing in diffusion models},
  author={Gandikota, Rohit and Orgad, Hadas and Belinkov, Yonatan and Materzy{\'n}ska, Joanna and Bau, David},
  booktitle={Proceedings of the IEEE/CVF winter conference on applications of computer vision},
  pages={5111--5120},
  year={2024}
}

@inproceedings{Lu2024MACEMC,
  title={Mace: Mass concept erasure in diffusion models},
  author={Lu, Shilin and Wang, Zilan and Li, Leyang and Liu, Yanzhu and Kong, Adams Wai-Kin},
  booktitle={Proceedings of the IEEE/CVF Conference on Computer Vision and Pattern Recognition},
  pages={6430--6440},
  year={2024}
}

@inproceedings{Gong2024ReliableAE,
  title={Reliable and efficient concept erasure of text-to-image diffusion models},
  author={Gong, Chao and Chen, Kai and Wei, Zhipeng and Chen, Jingjing and Jiang, Yu-Gang},
  booktitle={European Conference on Computer Vision},
  pages={73--88},
  year={2024},
  organization={Springer}
}

@article{Chin2023Prompting4DebuggingRT,
  title={Prompting4debugging: Red-teaming text-to-image diffusion models by finding problematic prompts},
  author={Chin, Zhi-Yi and Jiang, Chieh-Ming and Huang, Ching-Chun and Chen, Pin-Yu and Chiu, Wei-Chen},
  journal={arXiv preprint arXiv:2309.06135},
  year={2023}
}

@inproceedings{Tsai2023RingABellHR,
  title={Ring-a-bell! how reliable are concept removal methods for diffusion models?},
  author={Tsai, Yu-Lin and Hsu, Chia-Yi and Xie, Chulin and Lin, Chih-Hsun and Chen, Jia You and Li, Bo and Chen, Pin-Yu and Yu, Chia-Mu and Huang, Chun-Ying},
  booktitle={The Twelfth International Conference on Learning Representations},
  year={2023}
}

@inproceedings{Radford2021LearningTV,
  title={Learning transferable visual models from natural language supervision},
  author={Radford, Alec and Kim, Jong Wook and Hallacy, Chris and Ramesh, Aditya and Goh, Gabriel and Agarwal, Sandhini and Sastry, Girish and Askell, Amanda and Mishkin, Pamela and Clark, Jack and others},
  booktitle={International conference on machine learning},
  pages={8748--8763},
  year={2021},
  organization={PmLR}
}

@misc{LAION-AES,
    author={LAION-AI},
  title = {aesthetic-predictor},
  howpublished = {\url{https://github.com/LAION-AI/aesthetic-predictor}},
  year={2022}
}

@article{Meng2021SDEditGI,
  title={Sdedit: Guided image synthesis and editing with stochastic differential equations},
  author={Meng, Chenlin and He, Yutong and Song, Yang and Song, Jiaming and Wu, Jiajun and Zhu, Jun-Yan and Ermon, Stefano},
  journal={arXiv preprint arXiv:2108.01073},
  year={2021}
}

@article{Bui2024ErasingUC,
  title={Erasing undesirable concepts in diffusion models with adversarial preservation},
  author={Bui, Anh and Vuong, Long and Doan, Khanh and Le, Trung and Montague, Paul and Abraham, Tamas and Phung, Dinh},
  journal={arXiv preprint arXiv:2410.15618},
  year={2024}
}

@inproceedings{Fuchi2024ErasingCF,
  title={Erasing Concepts from Text-to-Image Diffusion Models with Few-shot Unlearning.},
  author={Fuchi, Masane and Takagi, Tomohiro},
  booktitle={BMVC},
  year={2024}
}

@inproceedings{Kumari2023AblatingCI,
  title={Ablating concepts in text-to-image diffusion models},
  author={Kumari, Nupur and Zhang, Bingliang and Wang, Sheng-Yu and Shechtman, Eli and Zhang, Richard and Zhu, Jun-Yan},
  booktitle={Proceedings of the IEEE/CVF international conference on computer vision},
  pages={22691--22702},
  year={2023}
}

@inproceedings{Gandikota2023ErasingCF,
  title={Erasing concepts from diffusion models},
  author={Gandikota, Rohit and Materzynska, Joanna and Fiotto-Kaufman, Jaden and Bau, David},
  booktitle={Proceedings of the IEEE/CVF international conference on computer vision},
  pages={2426--2436},
  year={2023}
}

@article{na2025trainingfree,
  title={Training-free safe text embedding guidance for text-to-image diffusion models},
  author={Na, Byeonghu and Kang, Mina and Kwak, Jiseok and Park, Minsang and Shin, Jiwoo and Jun, SeJoon and Lee, Gayoung and Kim, Jin-Hwa and Moon, Il-Chul},
  journal={Advances in Neural Information Processing Systems},
  volume={38},
  pages={85984--86014},
  year={2026}
}

@article{Achiam2023GPT4TR,
  title={Gpt-4 technical report},
  author={Achiam, Josh and Adler, Steven and Agarwal, Sandhini and Ahmad, Lama and Akkaya, Ilge and Aleman, Florencia Leoni and Almeida, Diogo and Altenschmidt, Janko and Altman, Sam and Anadkat, Shyamal and others},
  journal={arXiv preprint arXiv:2303.08774},
  year={2023}
}

@inproceedings{Lipman2022FlowMF,
  title={Flow matching for generative modeling},
  author={Lipman, Yaron and Chen, Ricky TQ and Ben-Hamu, Heli and Nickel, Maximilian and Le, Matthew},
  booktitle={The eleventh international conference on learning representations},
  year={2022}
}

@inproceedings{Liu2022FlowSA,
  title={Flow straight and fast: Learning to generate and transfer data with rectified flow},
  author={Liu, Xingchao and Gong, Chengyue and Liu, Qiang},
  booktitle={International conference on learning representations (ICLR)},
  year={2023}
}

@inproceedings{Peebles2022ScalableDM,
  title={Scalable diffusion models with transformers},
  author={Peebles, William and Xie, Saining},
  booktitle={Proceedings of the IEEE/CVF international conference on computer vision},
  pages={4195--4205},
  year={2023}
}

@inproceedings{Esser2024ScalingRF,
  title={Scaling rectified flow transformers for high-resolution image synthesis},
  author={Esser, Patrick and Kulal, Sumith and Blattmann, Andreas and Entezari, Rahim and M{\"u}ller, Jonas and Saini, Harry and Levi, Yam and Lorenz, Dominik and Sauer, Axel and Boesel, Frederic and others},
  booktitle={Forty-first international conference on machine learning},
  year={2024}
}

@misc{blackforestlabs2024,
    author = {Black Forest Labs},
    title = {FLUX},
    howpublished = {\url{https://blackforestlabs.ai/announcing-black-forest-labs/}},
    year = {2024},
    note = {Accessed: [19.11.2025]}
}

@article{Barez2025OpenPI,
  title={Open problems in machine unlearning for ai safety},
  author={Barez, Fazl and Fu, Tingchen and Prabhu, Ameya and Casper, Stephen and Sanyal, Amartya and Bibi, Adel and O'Gara, Aidan and Kirk, Robert and Bucknall, Ben and Fist, Tim and others},
  journal={arXiv preprint arXiv:2501.04952},
  year={2025}
}

@article{Wei2025ResponsibleDA,
  title={Responsible Diffusion: A Comprehensive Survey on Safety, Ethics, and Trust in Diffusion Models},
  author={Wei, Kang and Yuan, Xin and Huo, Fushuo and Ma, Chuan and Yuan, Long and Li, Songze and Ding, Ming and Tao, Dacheng},
  journal={arXiv preprint arXiv:2509.22723},
  year={2025}
}

@inproceedings{Li2025SetYS,
  title={Set you straight: Auto-steering denoising trajectories to sidestep unwanted concepts},
  author={Li, Leyang and Lu, Shilin and Ren, Yan and Kong, Adams Wai-Kin},
  booktitle={Proceedings of the 33rd ACM International Conference on Multimedia},
  pages={9257--9266},
  year={2025}
}

@inproceedings{Gao2024EraseAnythingEC,
  title={Eraseanything: Enabling concept erasure in rectified flow transformers},
  author={Gao, Daiheng and Lu, Shilin and Zhou, Wenbo and Chu, Jiaming and Zhang, Jie and Jia, Mengxi and Zhang, Bang and Fan, Zhaoxin and Zhang, Weiming},
  booktitle={Forty-second International Conference on Machine Learning},
  year={2025}
}

@inproceedings{Yoon2024SAFREETA,
  title={Safree: Training-free and adaptive guard for safe text-to-image and video generation},
  author={Yoon, Jaehong and Yu, Shoubin and Patil, Vaidehi Ramesh and Yao, Huaxiu and Bansal, Mohit},
  booktitle={International Conference on Learning Representations},
  volume={2025},
  pages={56439--56465},
  year={2025}
}

@article{Raffel2019ExploringTL,
  title={Exploring the limits of transfer learning with a unified text-to-text transformer},
  author={Raffel, Colin and Shazeer, Noam and Roberts, Adam and Lee, Katherine and Narang, Sharan and Matena, Michael and Zhou, Yanqi and Li, Wei and Liu, Peter J},
  journal={Journal of machine learning research},
  volume={21},
  number={140},
  pages={1--67},
  year={2020}
}

@article{Dathathri2019PlugAP,
  title={Plug and play language models: A simple approach to controlled text generation},
  author={Dathathri, Sumanth and Madotto, Andrea and Lan, Janice and Hung, Jane and Frank, Eric and Molino, Piero and Yosinski, Jason and Liu, Rosanne},
  journal={arXiv preprint arXiv:1912.02164},
  year={2019}
}

@inproceedings{Subramani2022ExtractingLS,
  title={Extracting latent steering vectors from pretrained language models},
  author={Subramani, Nishant and Suresh, Nivedita and Peters, Matthew E},
  booktitle={Findings of the Association for Computational Linguistics: ACL 2022},
  pages={566--581},
  year={2022}
}

@inproceedings{Rimsky2023SteeringL2,
  title={Steering llama 2 via contrastive activation addition},
  author={Rimsky, Nina and Gabrieli, Nick and Schulz, Julian and Tong, Meg and Hubinger, Evan and Turner, Alexander},
  booktitle={Proceedings of the 62nd Annual Meeting of the Association for Computational Linguistics (Volume 1: Long Papers)},
  pages={15504--15522},
  year={2024}
}

@article{Turner2023SteeringLM,
  title={Steering language models with activation engineering},
  author={Turner, Alexander Matt and Thiergart, Lisa and Leech, Gavin and Udell, David and Vazquez, Juan J and Mini, Ulisse and MacDiarmid, Monte},
  journal={arXiv preprint arXiv:2308.10248},
  year={2023}
}

@article{Zhao2026ODESteerAU,
  title={Odesteer: A unified ode-based steering framework for llm alignment},
  author={Zhao, Hongjue and Sun, Haosen and Kong, Jiangtao and Li, Xiaochang and Wang, Qineng and Jiang, Liwei and Zhu, Qi and Abdelzaher, Tarek and Choi, Yejin and Li, Manling and others},
  journal={arXiv preprint arXiv:2602.17560},
  year={2026}
}

@inproceedings{Gaintseva2025CASteerCS,
  title={CASteer: Cross-Attention Steering for Controllable Concept Erasure},
  author={Tatiana Gaintseva and Andreea-Maria Oncescu and Chengcheng Ma and Ziquan Liu and Martin Benning and Gregory G. Slabaugh and Jiankang Deng and Ismail Elezi},
  year={2025},
  url={https://api.semanticscholar.org/CorpusID:276961220}
}

@inproceedings{rodriguez2025controlling,
  title={Controlling language and diffusion models by transporting activations},
  author={Rodriguez, Pau and Blaas, Arno and Klein, Michal and Zappella, Luca and Apostoloff, Nicholas and Suau, Xavier and others},
  booktitle={International Conference on Learning Representations},
  volume={2025},
  pages={89812--89855},
  year={2025}
}

@article{DBLP:conf/icml/CywinskiD25,
  title={Saeuron: Interpretable concept unlearning in diffusion models with sparse autoencoders},
  author={Deja, Kamil and others},
  journal={arXiv preprint arXiv:2501.18052},
  year={2025}
}

@inproceedings{Ronneberger2015UNetCN,
  title={U-net: Convolutional networks for biomedical image segmentation},
  author={Ronneberger, Olaf and Fischer, Philipp and Brox, Thomas},
  booktitle={International Conference on Medical image computing and computer-assisted intervention},
  pages={234--241},
  year={2015},
  organization={Springer}
}

@article{Hu2021LoRALA,
  title={Lora: Low-rank adaptation of large language models.},
  author={Hu, Edward J and Shen, Yelong and Wallis, Phillip and Allen-Zhu, Zeyuan and Li, Yuanzhi and Wang, Shean and Wang, Liang and Chen, Weizhu and others},
  journal={Iclr},
  volume={1},
  number={2},
  pages={3},
  year={2022}
}

@inproceedings{Leong2023SelfDetoxifyingLM,
  title={Self-detoxifying language models via toxification reversal},
  author={Leong, Chak Tou and Cheng, Yi and Wang, Jiashuo and Wang, Jian and Li, Wenjie},
  booktitle={Proceedings of the 2023 Conference on Empirical Methods in Natural Language Processing},
  pages={4433--4449},
  year={2023}
}

@inproceedings{Wang2023TrojanAA,
  title={Trojan activation attack: Red-teaming large language models using steering vectors for safety-alignment},
  author={Wang, Haoran and Shu, Kai},
  booktitle={Proceedings of the 33rd ACM International Conference on Information and Knowledge Management},
  pages={2347--2357},
  year={2024}
}

@inproceedings{Yang2023MMADiffusionMA,
  title={Mma-diffusion: Multimodal attack on diffusion models},
  author={Yang, Yijun and Gao, Ruiyuan and Wang, Xiaosen and Ho, Tsung-Yi and Xu, Nan and Xu, Qiang},
  booktitle={Proceedings of the IEEE/CVF Conference on Computer Vision and Pattern Recognition},
  pages={7737--7746},
  year={2024}
}

@article{Gatys2015ANA,
  title={A neural algorithm of artistic style},
  author={Gatys, Leon A and Ecker, Alexander S and Bethge, Matthias},
  journal={arXiv preprint arXiv:1508.06576},
  year={2015}
}

@article{bedapudi2019nudenet,
  title={Nudenet: Neural nets for nudity classification, detection and selective censoring},
  author={Bedapudi, P},
  year={2019},
  publisher={December}
}

@misc{GIPHYDetector,
  title        = "Giphy celebrity detector.",
  author       = "{ Nick Hasty, Ihor Kroosh, Dmitry Voitekh, and Dmytro Korduban}",
  howpublished = "\url{https://github.com/Giphy/celeb-detection-oss.}",
  year         = 2019,
}

\end{document}